\documentclass{article}

    \PassOptionsToPackage{numbers, compress}{natbib}
\usepackage[final]{neurips_2025}

\usepackage[utf8]{inputenc} % allow utf-8 input
\usepackage[T1]{fontenc}    % use 8-bit T1 fonts
\usepackage{hyperref}       % hyperlinks
\usepackage{url}            % simple URL typesetting
\usepackage{booktabs}       % professional-quality tables
\usepackage{amsfonts}       % blackboard math symbols
\usepackage{nicefrac}       % compact symbols for 1/2, etc.
\usepackage{microtype}      % microtypography
\usepackage{xcolor}         % colors

\usepackage{amsmath}
\usepackage[pdftex]{graphicx}
\usepackage{xspace}
\usepackage[dvipsnames]{xcolor}
\usepackage{multirow}
\usepackage{multicol}
\usepackage{wrapfig}
\usepackage{makecell}

\usepackage{tcolorbox}       % 核心包
\tcbuselibrary{skins}        % 支持高级样式（如透明度）
\tcbuselibrary{breakable}    % 支持分页（若需跨页断行）
\usepackage{float}

\newcommand{\ourmethod}{DSAS\xspace}
\title{DSAS: A Universal Plug-and-Play Framework for Attention Optimization in Multi-Document Question Answering}
\definecolor{tab_green}{HTML}{81B21F}

\author{%
  Jiakai Li$^{1}$, Rongzheng Wang$^{1}$, Yizhuo Ma$^{1}$, \\
  \textbf{Shuang Liang}$^{1}$, \textbf{Guangchun Luo}$^{2}$, \textbf{Ke Qin}$^{1}\thanks{Corresponding author.}$ \\
  $^{1}$Institute of Intelligent Computing, \\ University of Electronic Science and Technology of China, Chengdu, China\\
  $^{2}$School of Information and Software Engineering, \\ University of Electronic Science and Technology of China, Chengdu, China\\
  \texttt{ljk@std.uestc.edu.cn, qinke@uestc.edu.cn}\\
}

\begin{document}

\maketitle

\begin{abstract}
  While large language models (LLMs) show considerable promise across various fields, they have notable limitations in handling multi-document question answering (Multi-doc QA) tasks. The first challenge is long-range dependency modeling, where LLMs struggle to focus on key information in long texts, which weakens important semantic connections. Second, most LLMs suffer from the ``lost-in-the-middle'' issue, where they have difficulty processing information in the middle of long inputs. Current solutions either truncate global dependencies or demand costly finetuning, ultimately lacking a universal and simple solution for these challenges.
  To resolve these limitations, we propose Dual-Stage Adaptive Sharpening (\ourmethod) containing two modules. (i) The Contextual Gate Weighting (CGW) module alleviates ``lost-in-the-middle'' by assessing paragraph relevance through layer-wise attention tracking and position-aware weighting. (ii) The Reciprocal Attention Suppression (RAS) module enhances focus on critical paragraphs by suppressing information exchange between key and irrelevant texts, thus mitigating the limitations in long-range dependency modeling. Notably, \ourmethod functions as a plug-and-play solution requiring no architectural modifications or extra training parameters.
  Extensive experiments on four benchmarks demonstrate DSAS's efficacy across mainstream LLMs (Llama, Qwen, Mistral, and Deepseek), with an average F1-score improvement of 4.2\% in Multi-doc QA tasks on Llama-3.1-8B-Instruct and Qwen2.5-14B-Instruct. Ablation studies confirm the essential contributions of both the CGW and RAS modules. In addition, detailed discussions in the Appendix further validate the robustness and scalability of \ourmethod.
\end{abstract}

\section{Introduction}
Transformer-based~\cite{attention} large language models (LLMs) have demonstrated remarkable performance, which have extensively promoted various complex natural language processing applications~\cite{graphrag,betalr,tog,dda,graphtool}. Building on the progress, recent advancements have shifted research focus toward enhancing LLMs' long-context processing capabilities~\cite{long_survey}, giving rise to LLMs that significantly expand context windows from 4K tokens to 128K or even 1M tokens, e.g., Llama-3.1-8B-Instruct~\cite{llama3} and Gemini-1.5~\cite{gemini}. These LLMs have unlocked unprecedented potential for complex tasks requiring cross-document reasoning, such as legal case analysis~\cite{legal_case} and multi-source academic synthesis~\cite{academic_case}. However, simply concatenating multiple documents into these long contexts often results in degraded performance due to attention dilution~\cite{attn_dilution}, a phenomenon where critical inter-document dependencies are overshadowed by irrelevant tokens. While solutions like StreamingLLM~\cite{StreamingLLM} and Selective Self-attention~\cite{SSA} are introduced, they either truncate global dependencies or lack generalizability, leaving the core challenge of context-aware attention prioritization unresolved.

\begin{figure}[t]
    \centering
    \includegraphics[width=\linewidth]{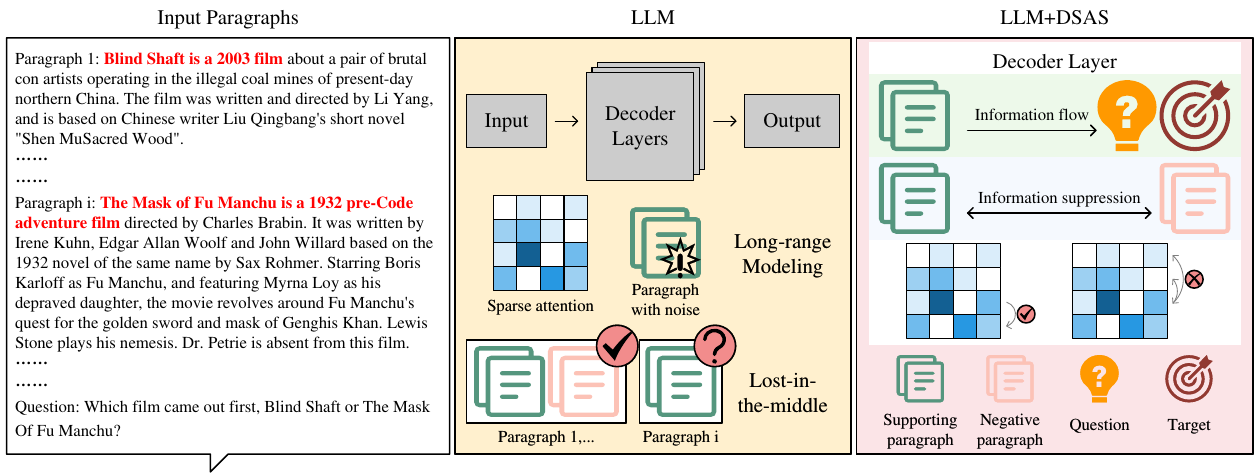}
    \caption{In Multi-doc QA tasks, directly processing long inputs comprising numerous paragraphs with LLMs presents two major challenges: long-range dependency modeling and ``lost-in-the-middle'', resulting in degraded answer quality. \ourmethod acts as a plug-in that enhances LLMs through a dual-stage process: (i) reinforcing information flow between supporting passages and both the question and target, and (ii) suppressing interactions between supporting and negative paragraphs.}
    \label{fig:intro}
\end{figure}

As shown in Figure~\ref{fig:intro}, the aforementioned attention dilution phenomenon reveals two critical limitations in multi-document question answering (Multi-doc QA) scenarios:
(i) Limited long-range dependency modeling: Despite claims of 128K-token support, RULER~\cite{ruler} reveals that LLMs often struggle with real-world tasks requiring combinatorial reasoning. While recent efforts (e.g., StreamingLLM~\cite{StreamingLLM}, LM-Infinite~\cite{lm_infinite}) explore attention mechanism optimization to address the challenge, they sacrifice global token interactions, compromising Multi-doc QA performance.
(ii) Persistent ``lost-in-the-middle'' issue: Nelson et al.~\cite{lost_in_the_middle} demonstrate that LLMs perform poorly when key information appears in the middle of long inputs. Current solutions like LongAlign~\cite{longalign} using a hybrid strategy (combining long-instruction examples with short data) require additional training using curated datasets, making them less adaptable to mainstream LLMs.
These challenges highlight the critical need for universal, plug-and-play modules that enhance Multi-doc QA capabilities without architectural constraints or task-specific fine-tuning.

Recent studies~\cite{label_anchor,instance_adaptive,opera} demonstrate that attention-driven information flow analysis provides critical insights into model reasoning patterns. Inspired by this insight, we adapt the methodology to Multi-doc QA tasks. To address these issues, we introduce Dual-Stage Adaptive Sharpening (\ourmethod) as shown in Figure~\ref{fig:intro}, a training-free attention optimization mechanism comprising two modules: Contextual Gate Weighting (CGW) and Reciprocal Attention Suppression (RAS). Specifically, our method works as follows: CGW tracks attention scores across selected model layers and calculates a contextual gate weight for each paragraph. Additionally, CGW introduces a position-aware weighting mechanism to enhance focus on information in the middle. Then RAS identifies critical paragraphs and attenuates information exchange between critical paragraphs and irrelevant content. \ourmethod requires no architectural changes or extra finetuning, serving as a universal plug-in for Transformer-based LLMs to strengthen Multi-doc QA capabilities.

Our contributions are as follows: 
(i) We systematically investigate the information flows on several LLMs in Multi-doc QA through paragraph disparity level and answer quality level analysis.
(ii) Based on the findings, we propose \ourmethod, a training-free universal plug-in for Transformer-based LLMs, which enhances focus on critical information through CGW and RAS, while suppressing irrelevant content. 
(iii) Extensive experiments on four public benchmark datasets demonstrate the effectiveness of \ourmethod for various LLMs including Llama, Qwen, Mistral and Deepseek, achieving an average F1-score improvement of up to 4.2\% on Llama-3.1-8B-Instruct~\cite{llama3} and Qwen2.5-14B-Instruct~\cite{qwen}.

\section{Information Flow Analysis on Multi-doc QA} \label{sec:2}
Identifying key factors for effective Multi-doc QA reasoning is essential. To achieve this, we conduct a systematic analysis of LLMs' inference processes across three core components: input paragraphs $(p)$, question $(q)$, and target $(t)$. A critical step lies in selecting suitable methods to study the semantic interactions among these components. Attention score analysis~\cite{attn_score_ana}, a method widely used to examine information flow, is adapted here to investigate how LLMs integrate cross-document information.

\subsection{Preliminaries}\label{sec:pre}
Let $A_{h,l}^S, A_{h,l}^W$ denote the attention score matrix and the attention weight matrix of the $h$-th head in the $l$-th layer, respectively. We obtain the layer-specific matrices $A_l^S$ and $A_l^W$ by summing across all attention heads. Here, $A_l^W(i,j)$ represents the information flow from the $j$-th to the $i$-th token.

To analyze interactions among three key components: (i) paragraphs ($p^1$, \ldots, $p^C$): Each paragraph $p^m$ spans input token indices \{$p^m_s$,$p^m_s+1$,\ldots,$p^m_e$\}, where $p^m_s$ and $p^m_e$ denote the start/end positions and $C$ is the total paragraph count; (ii) question $q$; and (iii) target $t$: the answer generation position (i.e., the final token in the input), we propose two metrics. $\mathcal{I}_{p^m,q}$ and $\mathcal{I}_{p^m,t}$ are the Top-K significance of the information flow from the $m$-th paragraph $p^m$ to $q$ and $p^m$ to $t$, respectively:
\begin{align}
    \mathcal{I}_{p^m,q} &= \frac{1}{Q}\sum \text{Top-K}\left( \left\{ \sum_{i \in q} A_l^W(i,j) | j \in p^m\right\}\right), \\
    \mathcal{I}_{p^m,t} &= \sum \text{Top-K}\left( \left\{ A_l^W(t,j) | j \in p^m\right\}\right), m \in \{1,\ldots,C\},
\end{align}
where $Q$ denote the token length of the question. $\mathcal{I}_{p^m,q}$ and $\mathcal{I}_{p^m,t}$ serve as information flow indicators to analyze the inference process for LLMs. The larger their values are, the more the generated results pay attention to the corresponding paragraphs.

\textbf{Experimental Settings.} We choose Llama-3.1-8B-Instruct~\cite{llama3} and Qwen2.5-7B-Instruct~\cite{qwen} as our primary models for investigation, due to their moderate model size and strong instruction following ability. For datasets, since LongBench~\cite{longbench} lacks supporting facts (labels for supporting paragraphs), we use HotpotQA~\cite{hotpotqa}, 2WikiMultiHopQA~\cite{2wikimqa} and MuSiQue~\cite{musique}. We sample 1000 examples from the training set for evaluation. Templates for constructing inputs are provided in Appendix~\ref{app:templ}. 
\begin{figure}[t]
    \centering
    \includegraphics[width=\textwidth]{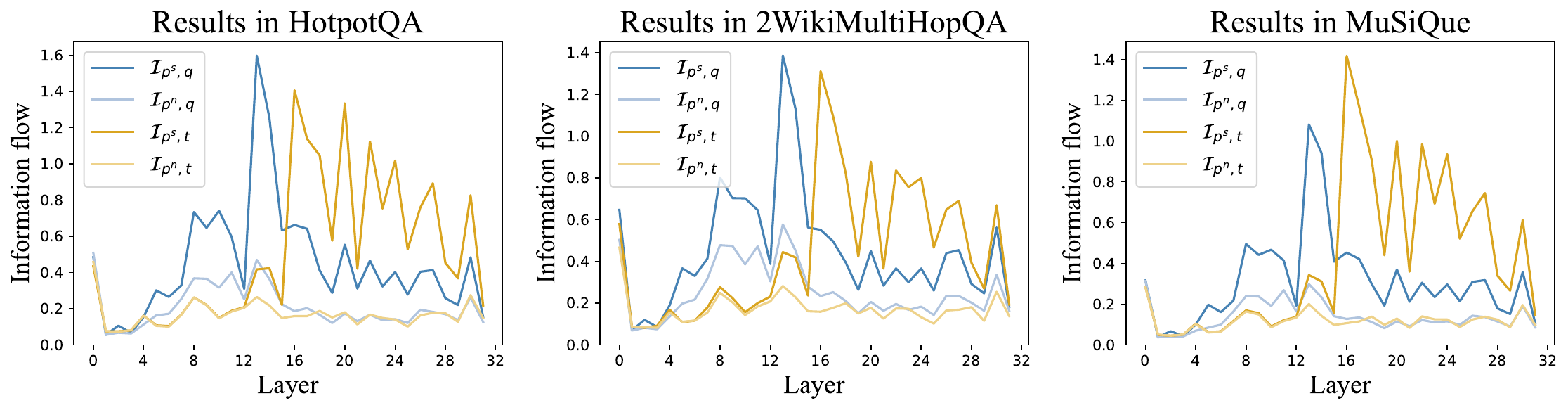}
    \vspace{-1em}
    \caption{Layer-wise information flows of HotpotQA, 2WikiMultiHopQA and MuSiQue tested on Llama-3.1-8B-Instruct. $p^s$ and $p^n$ denote supporting paragraphs and negative paragraphs, respectively. The results of Qwen2.5-7B-Instruct are shown in Appendix~\ref{app:para_ana}.}
    \label{fig:layer_ana}
\end{figure}

\subsection{Paragraph Disparity Level Analysis}\label{sec:lay_ana}
Most LLMs adopt a multi-layer Transformer architecture, with each decoder block processing semantic information differently during inference. We intend to check layer-wise information flows of $\mathcal{I}_{p^m,q}$ and $\mathcal{I}_{p^m,t}$. Figure~\ref{fig:layer_ana} visualizes layer-wise attention weight values. These values serve as a quantified metric to display how the different paragraphs contribute to the generated answer from the paragraph disparity perspective: Information flows from distinct paragraphs diverge only slightly in the shallow layers of the LLM, which suggests that the LLM first establishes basic semantic understanding to support deeper processing. As layers deepen, a clear divergence emerges between $\mathcal{I}_{p^s,q}$ and $\mathcal{I}_{p^n,q}$, demonstrating the LLM progressively distinguishes task-relevant paragraphs through its layer-wise semantic processing, enabling rational attention allocation. Ultimately, in the deep layers of the LLM, LLM recognizes and utilizes key paragraphs for answer generation. 

Our analysis uncovers a two-stage reasoning pattern: (i) Information initially converges on the question, with supporting paragraphs exhibiting stronger information flows than negative paragraphs, confirming the model's ability to prioritize semantically relevant content. (ii) Subsequently, the information from supporting paragraphs aggregates to the target, where the LLM strategically leverages critical paragraphs to formulate answers. These observations highlight the inherent interpretability of LLMs, proving their capabilities to integrate task-specific information during inference.

\subsection{Answer Quality Level Analysis}\label{sec:rea_ana}
\begin{wrapfigure}{r}{0.6\textwidth}
    \vspace{-1em}
    \centering
    \includegraphics[width=.6\textwidth]{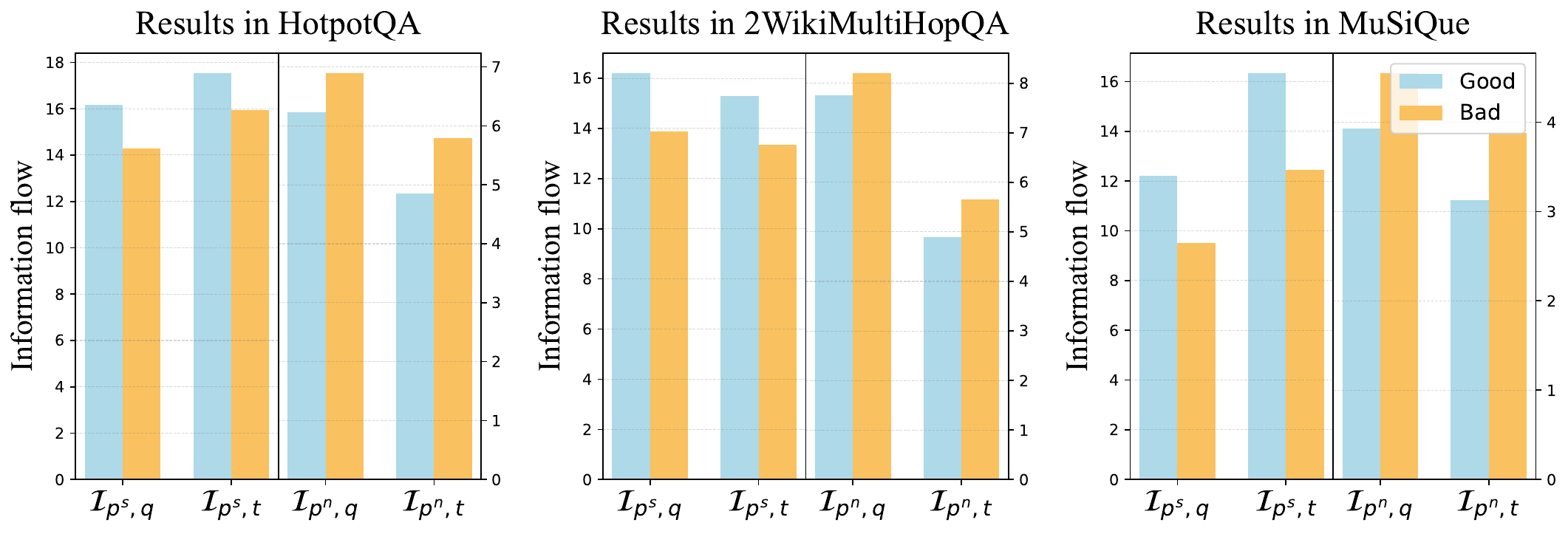}
    \vspace{-1.2em}
    \caption{Comparison between mean values of the good and bad instances from the 1000 samples of Llama-3.1-8B. The results of Qwen2.5-7B are shown in Appendix~\ref{app:ans_ana}.}
    \vspace{-1em}
    \label{fig:rea_ana}
\end{wrapfigure}
% \begin{figure}[h]
%     \centering
%     \includegraphics[width=\linewidth]{pics/bar_llama.pdf}
%     \caption{Comparison between mean values of the good and bad instances from the 1000 samples of Llama-3.1-8B. The results of Qwen2.5-7B are shown in Appendix~\ref{app:ans_ana}.}
%     \label{fig:rea_ana}
% \end{figure}
We define a reasoning process as \textsc{good} if its output exactly matches the reference answer, and \textsc{bad} if it contains no word-level overlap with the reference answer (see Appendix~\ref{app:ans_ass} for more details).
To investigate the divergence between good and bad reasoning patterns, we aggregate information flow values across all model layers and conduct a comparative analysis of two distinct groups: $\mathcal{I}_{p^s,q}$ and $\mathcal{I}_{p^s,t}$ from supporting paragraphs, along with $\mathcal{I}_{p^n,q}$ and $\mathcal{I}_{p^n,t}$ from negative paragraphs. As shown in Figure \ref{fig:rea_ana}, good reasoning exhibits higher values of $\mathcal{I}_{p^s,q}$ and $\mathcal{I}_{p^s,t}$, and lower values of $\mathcal{I}_{p^n,q}$ and $\mathcal{I}_{p^n,t}$. The MuSiQue dataset reveals the largest gaps between good and bad reasonings across all four metrics, which is probably attributable to its multi-hop reasoning complexity.

Notably, even in bad reasoning instances, $\mathcal{I}_{p^s,q}$ and $\mathcal{I}_{p^s,t}$ for supporting paragraphs consistently exceed those of $\mathcal{I}_{p^n,q}$ and $\mathcal{I}_{p^n,t}$ for negative paragraphs. This observation indicates that LLMs possess inherent discrimination capabilities between supporting and negative paragraphs regardless of reasoning quality, though the information flows of supporting paragraphs remain insufficient for optimal answer generation. Our framework addresses this core issue by introducing a quantitative information flow analysis that measures $\mathcal{I}_{p,q}$ and $\mathcal{I}_{p,t}$ for each paragraph. This analysis enables the selective and precise amplification of relevant semantic information, as detailed in Section \ref{sec:method}.

\section{Methodology} \label{sec:method}
\begin{figure}[t]
    \centering
    \includegraphics[width=\linewidth]{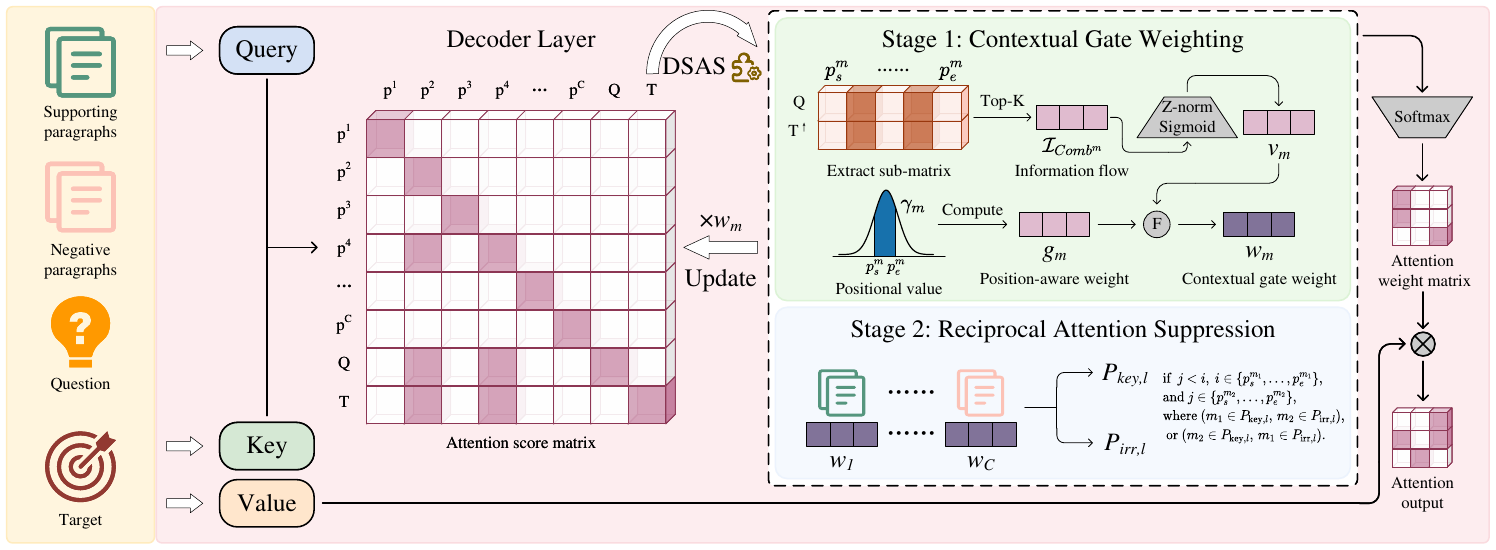}
    \caption{Illustration of Dual-Stage Adaptive Sharpening (\ourmethod), including Contextual Gate Weighting (CGW) and Reciprocal Attention Suppression (RAS).}
    \label{fig:method}
\end{figure}

To explore the Multi-doc QA capabilities of LLMs, we analyze the attention score matrix in each layer and identify the key paragraphs through the information flows. 
Building on the above findings, we propose Dual-Stage Adaptive Sharpening (\ourmethod), which consists of two key modules. In Section \ref{sec:CGW}, we introduce Contextual Gate Weighting (CGW), which identifies key paragraphs and strengthens the attention of the question and target position toward these texts. Section \ref{sec:RAS} describes Reciprocal Attention Suppression (RAS), which aims to suppress interactions between key paragraphs and irrelevant information.
At a high level, we strategically integrate CGW and RAS modules into the computation of multi-head attention to implement \ourmethod:
\begin{equation}
    \begin{aligned}
        A_{h,l}^S&=\frac{QW_{h,l}^Q(KW_{h,l}^K)^T}{\sqrt{d_k}},\quad A_l^S=\text{Stack}(A_{1,l}^S,\ldots,A_{H,l}^S),\\
        A_{h,l}^S&=\text{RAS}(\text{CGW}(A_{h,l}^S,A_l^S)),\\
        A_{h,l}^W&=\text{Softmax}(A_{h,l}^S,\text{dim}=-1),\\
        O_{h,l}&=A_{h,l}^W(VW_{h,l}^V),\quad O_l=\text{Concat}(O_{1,l},\ldots,O_{H,l})W_l^O,
    \end{aligned}
    \label{eq:CGW_RAS}
\end{equation}
where $W_{h,l}^Q, W_{h,l}^K, W_{h,l}^V$ are projection matrices of the $h$-th head in the $l$-th layer, $W_l^O$ is the output projection matrix. We \textsc{Stack} tensors along the first dim and \textsc{Concat} them along the last dim. The framework of \ourmethod is shown in Figure~\ref{fig:method}. To avoid ambiguity, descriptions of key symbols appearing in the following text are provided in Table~\ref{tab:symbol} in Appendix~\ref{app:symb}.
\subsection{Stage 1: Contextual Gate Weighting (CGW)} \label{sec:CGW}
For each selected layer, we compute the combined information flows of each paragraph using the attention score matrix $A_{l}^S$. Specifically, we extract the attention sub-matrix corresponding to the positions of each paragraph, question and target as follows:
\begin{equation}
M \in \mathbb{R}^{2Q \times p^m} = \underset{\text{Question Matrix}}{\underbrace{[A_l^S(q,p^m)]}} 
\Vert \ 
\underset{\text{Target Matrix}}{\underbrace{[A_l^S(t,p^m)\uparrow^{(Q)}]}},  
\end{equation}
where $Q, p^m$ denote the token length of the question and the $m$-th paragraph, respectively, and $\uparrow^{(Q)}$ represents the expand operation to match the question length.
The combined information flow for the $m$-th paragraph is then calculated by averaging the column-wise Top-K values of $M$: 
\begin{equation}
\mathcal{I}_{Comb^m} = \frac{1}{K}\sum\text{Top-K} \left[ \sum_{i=1}^{2Q} M_{i,j} \right]_{j=1}^{p^m}, m \in \{1,\ldots,C\}.
\label{eq:top_k}
\end{equation}
%% Importance-aware weighting
We then compute $v_m$ through Z-normalization and sigmoid scaling:
\begin{equation}
v_m = 0.5\cdot\sigma\left( \frac{\mathcal{I}_{Comb^m} - \mu_I}{\sigma_I} \right)+0.5,m \in \{1,\ldots,C\}
\end{equation}
where $\sigma(\cdot)$ is the sigmoid function and $\mu_I,\sigma_I$ are the mean and the standard deviation of $\mathcal{I}_{Comb}$. The minimum value of $v_m$ is set to 0.5 to prevent overlooking the paragraphs excessively.
%% Position-Importance weighting
In addition, to mitigate the ``lost-in-the-middle'' issue, we introduce position-aware weighting that assigns greater weights to central key paragraphs. Hsieh et al.~\cite{ushape} attribute this limitation to the inherent U-shaped attention bias in LLMs, which places more emphasis on content at the beginning and end of the sequence. We correct this bias through the probability density function (PDF) of a Gaussian distribution, which is defined as follows:
\begin{equation}
    f(x) = \frac{1}{\sigma_p\sqrt{2\pi}} \cdot \exp\left(-\frac{(x-\mu_p)^2}{2\sigma_p^2}\right), \quad F(x)=\int_{-\infty}^{x} f(t) \, \mathrm{d}t
\end{equation}
where $\mu_p,\sigma_p$ are the mean and the standard deviation of the input token indices $\{0,1,\ldots,L-1\}$ (i.e., $\mu_p$ equals $0.5\cdot (L-1)$, $\sigma_p$ equals $\sqrt{\frac{L^2-1}{12}}$), and $F(x)$ is the cumulative distribution function (CDF) of a Gaussian distribution. Then the positional value $\gamma_m$ for the $m$-th paragraph is computed based on the token indices \{$p^m_s$,$p^m_s+1$,\ldots,$p^m_e$\} within the segment:
\begin{align}
    z_1&=\frac{p^m_s-\mu_p}{\sigma_p}, \quad z_2=\frac{p^m_e-\mu_p}{\sigma_p}, \\
    \gamma_m&=\frac{F(z_2)-F(z_1)}{z_2-z_1},
\end{align}
where $z_1,z_2$ denote the normalized value of $p^m_s$ and $p^m_e$, respectively. Next, we rank paragraphs by $v_m$, compute position-aware weights for the top 50\% paragraphs, and assign a value of 1 to the rest, aiming to prevent the model from attracting excessive attention to non-critical middle paragraphs. The position-aware weight $g_m$ is calculated as follows:
\begin{equation}
    g_m = \left\{ 
\begin{array}{ll}
\left(\frac{0.5C+1}{\text{rank}_m}\right)^{\gamma_m}, & \text{if } \text{rank}_m \le 0.5C, \\
1, & \text{otherwise.}
\end{array}
\right.
\label{eq:gm}
\end{equation}
The final contextual gate weight $w_m$ is derived as:
\begin{align}
    w_m^{'}&=v_m \cdot {g_m}^\alpha, \label{eq:alpha}\\
    w_m&=(1-\beta)\frac{w_m^{'}-\text{min}(w_i^{'})}{\text{max}(w_i^{'})-\text{min}(w_i^{'})}+\beta,i \in \{1,\ldots,C\} \label{eq:beta}
\end{align}
where hyperparameters $\alpha$ and $\beta$ balance positional and content relevance and determine the minimum value of $w_m$, respectively. The attention score matrix $A_{h,l}^S$ is dynamically adjusted by applying contextual gate weights of each paragraph, thereby refining the model's focus based on their importance:
\begin{equation}
    A_{h,l}^S(i,j)=w_m \cdot A_{h,l}^S(i,j),\quad \text{if }i\in\{q,t\},j\in\{p_s^m,p_s^m+1,\ldots,p_e^m\} 
\end{equation}

\subsection{Stage 2: Reciprocal Attention Suppression (RAS)} \label{sec:RAS}
The CGW module computes contextual gate weights for each paragraph to evaluate their relevance to answer generation, enabling clear distinction between critical and irrelevant paragraphs. The RAS module aims to suppress interactions between non-critical and key paragraphs. To achieve this, we classify paragraphs using a threshold-based method: paragraphs with contextual gate weights higher than the mean value of all $w_m$ are identified as key paragraphs, while those below the threshold are labeled as irrelevant paragraphs. The key and irrelevant paragraphs are denoted as $P_{key,l}$ and $P_{irr,l}$ in the $l$-th layer. 
Next, reciprocal attention suppression is applied between these categories. This process adjusts the attention score matrix $A_{h,l}^S$ by suppressing values between key and irrelevant paragraphs. The reciprocal suppression is bidirectional, affecting both interactions between $P_{key,l}$ and $P_{irr,l}$ to break cross-paragraph interference. Formally, the suppression is implemented as:
\begin{equation}
    \begin{aligned}
        &A_{h,l}^S(i,j)=\min(w_{m_1}, w_{m_2}) \cdot A_{h,l}^S(i,j),\\
        &\text{if } \ j < i,\ i \in \{p_s^{m_1}, p_s^{m_1}{+}1, \ldots, p_e^{m_1}\}, \text{and }
        j \in \{p_s^{m_2}, p_s^{m_2}{+}1, \ldots, p_e^{m_2}\}, \\
        &\text{where } (m_1 \in P_{\text{key},l},\, m_2 \in P_{\text{irr},l}) \text{ or} \ (m_2 \in P_{\text{key},l},\, m_1 \in P_{\text{irr},l}).
    \end{aligned}
\end{equation}

\section{Experiments} \label{sec:exp}
This section assesses \ourmethod, a plug-and-play attention mechanism designed to enhance Multi-doc QA performance across diverse models and downstream tasks. Our experiments are structured as follows: Section \ref{sec:imp} details the implementation of \ourmethod, covering datasets, metrics, models and hyperparameter settings. In Section \ref{sec:main_res}, we show that \ourmethod achieves improvements on all Multi-doc QA benchmarks, including HotpotQA~\cite{hotpotqa}, 2WikiMultiHopQA~\cite{2wikimqa}, MuSiQue~\cite{musique}, and LongBench~\cite{longbench}, without requiring any additional training. Section \ref{sec:abla} presents ablation studies to examine the effectiveness of \ourmethod under different variants and hyperparameters. Section \ref{sec:further} further explores the robustness of \ourmethod.
\subsection{Implementation} \label{sec:imp}
% \begin{wraptable}{r}{.4\textwidth}
%  \vspace{-1.7em}
%  \centering
%  \caption{Statistics of the datasets.}
%  \vspace{1em}
%  \label{tab:stat_data}
%  \small
%     \begin{tabular}{ll}
%         \toprule
%         Datasets & Number \\ \midrule 
%         HotpotQA & 7,405 \\
%         2WikiMultiHopQA & 12,576 \\
%         MuSiQue & 2,417 \\
%         L-HotpotQA & 200 \\
%         L-2WikiMultiHopQA & 200 \\
%         L-MuSiQue & 200 \\
%         \bottomrule
%     \end{tabular}
% \end{wraptable}

\textbf{Datasets \& Metrics.} We evaluate our method on three classic Multi-doc QA benchmarks and a long-context dataset. Following previous works, we utilize the validation splits of HotpotQA~\cite{hotpotqa}, 2WikiMultiHopQA~\cite{2wikimqa}, and MuSiQue~\cite{musique} to test our effectiveness, as they provide reference answers. To further assess generalization, we extend experiments to LongBench's~\cite{longbench} corresponding subsets on Multi-doc QA tasks. Referring to HotpotQA and LongBench, we consider the F1-score as the evaluation metric for these datasets. The metrics details are shown in Appendix~\ref{app:ans_ass}.

\textbf{Baselines \& Models.} PINE~\cite{pine} analyzes attention patterns to re-assign input paragraph positions. Given its similar operation and target to \ourmethod, we include PINE as a baseline for comparison, alongside a vanilla LLM baseline that performs inference without altering the original pipeline. We select six popular LLMs, including Llama-3.2-3B-Instruct~\cite{llama3}, Mistral-7B-Instruct-v0.2~\cite{mistral}, Qwen2.5-7B-Instruct~\cite{qwen}, DeepSeek-R1-Distill-Llama-8B~\cite{deepseek}, Llama-3.1-8B-Instruct~\cite{llama3}, Qwen2.5-14B-Instruct~\cite{qwen}, and Qwen2.5-32B-Instruct~\cite{qwen}, since they are popular Transformer-based decoder-only LLMs, which are convenient for exploiting and analyzing inside architectures. All models are deployed with the ``bfloat16'' data format due to the balance between efficiency and performance. We set the generation mode to greedy-decoding for all methods with deterministic sampling parameters: do\_sample=False, temperature=0, top\_p=1, max\_new\_tokens=32 (follow the setting in LongBench~\cite{longbench}). This setting minimizes the impact of irrelevant confounders during inference, thereby ensuring that identical models and inputs always produce the same answers to fixed questions. All experiments are implemented using PyTorch 2.2.1, and executed on the CPU of two 32-core Intel(R) Xeon(R) @ 2.80GHz and GPU of 8$\times$ NVIDIA A800. The codebase is compatible with Python 3.12, and computations were accelerated using CUDA 12.1. 

\textbf{Hyperparameters.} The hyperparameters in Equations~\ref{eq:top_k},~\ref{eq:alpha}, and~\ref{eq:beta} are set to $K$ = 10, $\alpha$ = 1, and $\beta$ = 0.7 for all benchmarks and LLMs. \ourmethod is applied to the final 50\% layers of all LLMs.

\subsection{Main Results} \label{sec:main_res}
\begin{table*}[t]
    %TODO: PINE results
    \caption{Comparsion results on HotpotQA, 2WikiMultiHopQA, MuSiQue and LongBench benchmarks. 2WikiMQA denotes 2WikiMultiHopQA. The evaluation for all tasks is assessed through the F1-score (\%).}
    \label{tab:main_res}
    \resizebox{\textwidth}{!}{
    \begin{tabular}{c|c|cccccc|c} \toprule
    \multirow{2}{*}{Models} & \multirow{2}{*}{Methods} & \multirow{2}{*}{HotpotQA} & \multirow{2}{*}{2WikiMQA} & \multirow{2}{*}{MuSiQue} & \multicolumn{3}{c|}{LongBench} & \multirow{2}{*}{Average} \\ \cmidrule{6-8} 
    & & & & & HotpotQA & 2WikiMQA & MuSiQue & \\ \midrule
    \multirow{3}{*}{Llama-3.2-3B} 
    & Vanilla & 39.1 & 39.4 & 20.8 & 41.9 & 34.1 & 12.7 & 31.3\\
    & PINE & 39.0 & 39.6 & \bf 21.3 & 42.4 & 35.3 & 12.5 & 31.6\\
    & \ourmethod & \bf 40.6 \textcolor{tab_green}{(+1.5)} & \bf 39.7 \textcolor{tab_green}{(+0.3)} & 21.2 \textcolor{tab_green}{(+0.4)} & \bf 42.7 \textcolor{tab_green}{(+0.8)} & \bf 35.9 \textcolor{tab_green}{(+1.8)} & \bf 13.2 \textcolor{tab_green}{(+0.5)} & \bf 32.2 \textcolor{tab_green}{(+0.9)} \\ \midrule
    \multirow{3}{*}{Mistral-7B} 
    & Vanilla & 32.0 & 19.0 & 24.3 & 39.3 & 16.6 & 22.2 & 25.6\\
    & PINE & 32.4 & 19.5 & 24.6 & 40.2 & \bf16.9 & 24.1 & 26.3\\
    & \ourmethod & \bf 33.8 \textcolor{tab_green}{(+1.8)} & \bf 20.2 \textcolor{tab_green}{(+1.2)} & \bf 26.8 \textcolor{tab_green}{(+2.5)} & \bf 41.9 \textcolor{tab_green}{(+2.6)} & 16.6 \textcolor{tab_green}{(+0)} & \bf 25.3 \textcolor{tab_green}{(+3.1)} & \bf 27.4 \textcolor{tab_green}{(+1.8)} \\ \midrule
    \multirow{3}{*}{Qwen2.5-7B} 
    & Vanilla & 42.3 & 46.0 & 30.5 & 55.1 & 50.3 & 28.1 & 42.1 \\
    & PINE & 44.5 & 47.2 & 31.4 & 57.3 & 52.3 & 30.4 & 43.9\\
    & \ourmethod & \bf 46.1 \textcolor{tab_green}{(+3.8)} & \bf 49.9 \textcolor{tab_green}{(+3.9)} & \bf 35.0 \textcolor{tab_green}{(+4.5)}& \bf 57.7 \textcolor{tab_green}{(+2.6)} & \bf 52.7 \textcolor{tab_green}{(+2.4)} & \bf 33.4 \textcolor{tab_green}{(+5.3)} & \bf 45.8 \textcolor {tab_green}{(+3.7)} \\ \midrule
    \multirow{3}{*}{DeepSeek-R1-8B} 
    & Vanilla & 37.2 & 22.9 & 26.8 & 45.1 & 20.3 & 24.8 & 29.5 \\
    & PINE & 36.8 & 22.6 & 26.5 & 45.2 & 20.1 & 23.9 & 29.2\\
    & \ourmethod & \bf 39.0 \textcolor{tab_green}{(+1.8)} & \bf 23.4 \textcolor{tab_green}{(+0.5)} & \bf 28.0 \textcolor{tab_green}{(+1.2)}& \bf 47.2 \textcolor{tab_green}{(+2.1)} & \bf 21.2 \textcolor{tab_green}{(+0.8)} & \bf 26.7 \textcolor{tab_green}{(+1.9)} & \bf 30.9 \textcolor{tab_green}{(+1.4)} \\ \midrule
    \multirow{3}{*}{Llama-3.1-8B} 
    & Vanilla & 43.6 & 47.3 & 34.6 & 53.3 & 42.6 & 25.4  & 41.1 \\
    & PINE & 46.6 & 49.2 & 37.0 & 54.4 & 46.9 & 30.6 & 44.1\\
    & \ourmethod & \bf 47.1 \textcolor{tab_green}{(+3.5)} & \bf 50.8 \textcolor{tab_green}{(+3.5)} & \bf 39.2 \textcolor{tab_green}{(+4.6)} & \bf 56.5 \textcolor{tab_green}{(+3.2)} & \bf 47.3 \textcolor{tab_green}{(+4.7)} & \bf 32.0 \textcolor{tab_green}{(+6.6)} &  \bf 45.5 \textcolor{tab_green}{(+4.2)} \\ \midrule
    \multirow{3}{*}{Qwen2.5-14B} 
    & Vanilla & 48.2 & 55.3 & 38.0 & 57.6 & 53.1 & 32.8  & 47.5\\
    & PINE & 50.4 & 57.0 & 41.8 & 59.8 & 56.4 & 36.7 & 50.4\\
    & \ourmethod & \bf 51.8 \textcolor{tab_green}{(+3.6)} & \bf 58.2 \textcolor{tab_green}{(+2.9)} & \bf 43.8 \textcolor{tab_green}{(+5.8)} & \bf 60.9 \textcolor{tab_green}{(+3.3)} & \bf 56.1 \textcolor{tab_green}{(+3.0)} & \bf 39.3 \textcolor{tab_green}{(+6.5)} & \bf 51.7 \textcolor{tab_green}{(+4.2)} \\ \midrule 
    \multirow{3}{*}{Qwen2.5-32B} 
    & Vanilla & 48.8 & 60.7 & 42.3 & 58.6 & 48.2 & 35.0 & 48.9\\
    & PINE & 50.8 & 61.0 & 44.5 & 58.1 & 47.9 & 36.4 & 49.8\\
    & \ourmethod & \bf 50.8 \textcolor{tab_green}{(+2.0)} & \bf 62.2 \textcolor{tab_green}{(+1.5)} & \bf 45.4 \textcolor{tab_green}{(+3.1)} & \bf 59.5 \textcolor{tab_green}{(+0.9)} & \bf 50.5 \textcolor{tab_green}{(+2.3)} & \bf 39.9 \textcolor{tab_green}{(+4.9)} & \bf 51.4 \textcolor{tab_green}{(+2.5)} \\  
    \bottomrule
    \end{tabular}}
\end{table*}
Table \ref{tab:main_res} presents the main experimental results, revealing three principal insights:
(i) Simply replacing the original attention module with our \ourmethod enhances LLMs' performance across all Multi-doc QA benchmarks without requiring additional training, achieving average F1-score gains between 0.9\% and 4.2\%. Llama-3.1-8B and Qwen2.5-14B achieve the most significant improvements, particularly on LongBench tasks. These results demonstrate that \ourmethod acts as a catalyst to enhance Multi-doc QA performance across diverse model architectures and task configurations. In contrast, although PINE~\cite{pine} achieves performance gains under certain configurations, its improvements are not consistently observed.
(ii) Performance improvements hold consistently across model scales ranging from 3B to 32B parameters, with medium-sized LLMs generally exhibiting greater performance gains (e.g., maximum improvement of Llama-3.1-8B and Qwen2.5-14B, followed by Qwen2.5-7B). We attribute the pattern to two main reasons. (a) Medium-sized LLMs (e.g., Llama-3.1-8B, Qwen2.5-14B) demonstrate adequate semantic understanding yet remain vulnerable to input noise, especially in long-context scenarios. \ourmethod overcomes this by analyzing information flows to precisely identify critical paragraphs and amplify model focus. In contrast, smaller models (e.g., Llama-3.2-3B) lack sufficient comprehension capacity, while larger models (e.g., Qwen2.5-32B) approach the task performance ceiling, leaving little room for further gains. (b) Medium-sized LLMs exhibit greater vulnerability to the ``lost-in-the-middle'' phenomenon, while larger models are generally more robust to this issue. Overall, \ourmethod effectively unlocks the latent capabilities of diverse LLMs on Multi-doc QA tasks, regardless of their architecture or scale. 
% This trend likely arises from our method leveraging the LLM's internal information flows, which require precise comprehension of input content (a capability that scales with model size). Notably, while \ourmethod yields measurable improvements for Qwen2.5-32B, the gains are comparatively limited to middle-sized LLMs (7B, 8B, 14B). This could indicate that larger models approach performance ceilings on current task metrics. However, \ourmethod still effectively unlocks the latent capabilities of LLMs on Multi-doc QA tasks. 
(iii) The improvement varies across different benchmarks. For medium-sized LLMs (parameters 7B, 8B, and 14B), greater improvements emerge on MuSiQue (including its LongBench extensions) compared to other benchmarks. This discrepancy suggests that \ourmethod particularly enhances LLMs' performance on complex tasks, where their inherent Multi-doc QA abilities can be better activated. The performance gap between HotpotQA and L-HotpotQA of these LLMs may be attributed to two factors. (a) Ansong et al.~\cite{hotpot_error} identify annotation inconsistencies in HotpotQA that could result in unreliable assessments. (b) HotpotQA and L-HotpotQA only involve two-hop reasoning tasks with relatively lower complexity which are easy for these LLMs.

\subsection{Ablation Studies} \label{sec:abla}
In this section, we conduct ablation studies to evaluate the effectiveness of each component in \ourmethod. Table \ref{tab:abla} summarizes the performance of different variants on all tasks. 

\textbf{w/o CGW.} Applying contextual gate weight $w_m$ only in RAS to regulate paragraph interactions degrades all metrics. This indicates that enhancing information propagation from key paragraphs to both the question and the target is essential for answer generation.
\begin{table*}[t]
    \caption{Ablation study of \ourmethod on HotpotQA, 2WikiMultiHopQA, MuSiQue and LongBench benchmarks. 2WikiMQA denotes 2WikiMultiHopQA. p-a denotes position-aware weight $g_m$ in Equation~\eqref{eq:gm}. The evaluation for all tasks is assessed through the F1-score (\%).}
    \label{tab:abla}
    \resizebox{\textwidth}{!}{
    \begin{tabular}{c|c|cccccc|c} \toprule
    \multirow{2}{*}{Models} & \multirow{2}{*}{Variants} & \multirow{2}{*}{HotpotQA} & \multirow{2}{*}{2WikiMQA} & \multirow{2}{*}{MuSiQue} & \multicolumn{3}{c|}{LongBench} & \multirow{2}{*}{Average} \\ \cmidrule{6-8} 
    & & & & & HotpotQA & 2WikiMQA & MuSiQue & \\ \midrule
    \multirow{4}{*}{Llama-3.2-3B} 
    & \ourmethod & \bf 40.6 & \bf 39.7 & 21.2 & 42.7 & \bf 35.9 & \bf 13.2  & \bf 32.2 \\
    & w/o CGW & 40.6 & 39.5 & 20.4 & \bf 43.5 & 34.6 & 12.9  & 31.9 \\
    & w/o RAS & 39.7 & 39.0 & \bf 21.7 & 42.9 & 34.6 & 11.4  & 31.6 \\
    & w/o p-a & 39.1 & 36.8 & 19.0 & 42.4 & 34.5 & 10.6  & 30.4 \\ \midrule
    \multirow{4}{*}{Qwen2.5-7B} 
    & \ourmethod & \bf 46.1 & 49.9 & \bf 35.0 & \bf 57.7 & \bf 52.7 & \bf 33.4 & \bf 45.8 \\
    & w/o CGW & 45.5 & 48.2 & 33.8 & 56.6 & 50.3 & 31.1 & 44.3 \\
    & w/o RAS & 45.4 & 47.8 & 32.6 & 56.7 & 51.3 & 32.0 & 44.3 \\
    & w/o p-a & 44.7 & \bf 50.5 & 33.2 & 56.2 & 49.6 & 30.6 & 44.1 \\
    \midrule
    \multirow{4}{*}{Llama-3.1-8B} 
    & \ourmethod & \bf 47.1 & \bf 50.8 & \bf 39.2 & \bf 56.5 & \bf 47.3 & \bf 32.0  & \bf 45.5 \\
    & w/o CGW & 46.6 & 48.9 & 38.4 & 55.6 & 46.6 & 30.9  & 44.5 \\
    & w/o RAS & 45.7 & 49.0 & 38.7 & 54.9 & 46.8 & 31.4  & 44.4 \\
    & w/o p-a & 45.9 & 49.7 & 37.8 & 56.2 & 46.2 & 30.4 & 44.4 \\ \midrule
    \multirow{4}{*}{Qwen2.5-14B} 
    & \ourmethod & \bf 51.8 & 58.2 & \bf 43.8 & 60.9 & \bf 56.1 & \bf 39.3 & \bf 51.7 \\
    & w/o CGW & 49.5 & 57.2 & 40.8 & 58.6 & 55.3 & 38.2 & 50.0 \\
    & w/o RAS & 50.2 & 57.6 & 41.4 & 59.9 & 54.1 & 38.6 & 50.3 \\
    & w/o p-a & 51.3 & \bf 58.9 & 43.1 & \bf 61.2 & 55.6 & 39.2 & 51.6 \\
    \bottomrule
    \end{tabular}}
\end{table*}

\textbf{w/o RAS.} Results reveal that removing $w_m$ during information aggregation lowers LLMs' performance, suggesting that the absence of RAS allows irrelevant content in $P_{irr,l}$ to introduce noise into the semantics of key paragraphs $P_{key,l}$.

\textbf{w/o p-a weight.} Setting $\alpha$ = 0 in Equation \eqref{eq:alpha} weakens resistance to the ``lost-in-the-middle'' issue, confirming that adaptively weighting middle paragraphs through $g_m$ enhances answer quality on most benchmarks. Notably, Qwen2.5-14B with this configuration achieves competitive results relative to \ourmethod, indicating larger LLMs inherently mitigate ``lost-in-the-middle'' more effectively, which aligns with the observations in LongPiBench~\cite{longpibench}. Nevertheless, the position-aware weight generally improves performance on all benchmarks and LLMs, mitigating ``lost-in-the-middle'' issue.

\begin{figure}[t]
\begin{minipage}{\linewidth}
\centering
\includegraphics[width=\linewidth]{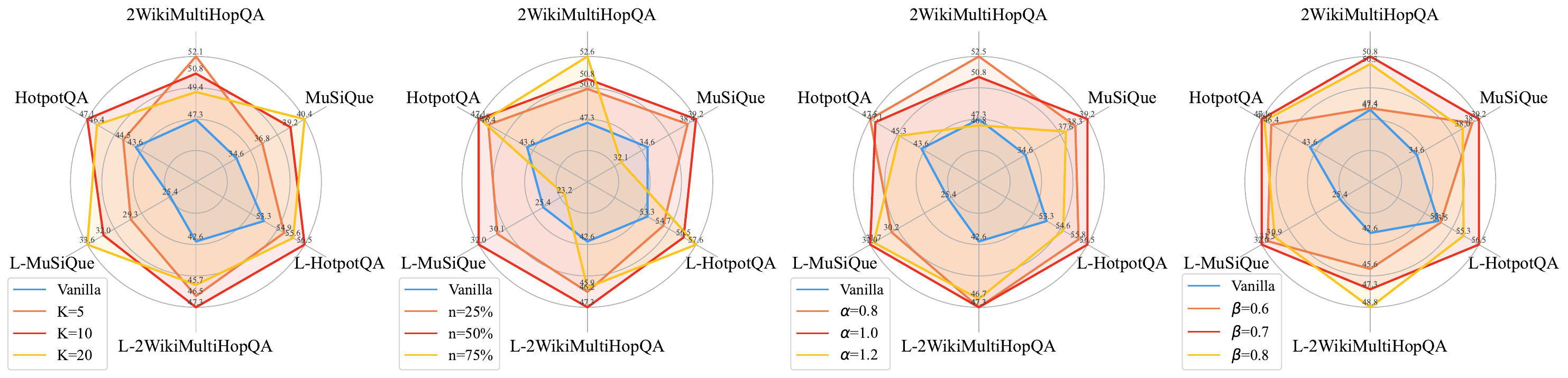}
\end{minipage}
\vfill
\centering
\vspace{-0.2em}
\includegraphics[width=\linewidth]{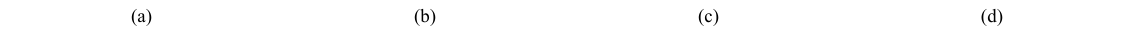}
\vspace{-1em}
\caption{Llama-3.1-8B's hyperparameter study of $K$, $n$, $\alpha$, $\beta$ on HotpotQA, 2WikiMultiHopQA, MuSiQue, and LongBench benchmarks. The hyperparameter studies of Llama-3.2-3B, Qwen2.5-7B, and Qwen2.5-14B are shown in Appendix~\ref{app:hyper_full}.}
\label{fig:hypers_llama8}
\end{figure}

\textbf{Hyperparameter Study.} 
The parameter $K$ in Equation~\eqref{eq:top_k} controls the number of tokens that the question and target attend to within each paragraph. Lower $K$ values risk insufficient contextual focus on the paragraphs, while higher values may degrade performance due to interference from irrelevant content. We evaluate $K$ with 5,10,20 and observe that $K$ = 10 achieves optimal performance of Llama-3.1-8B, as shown in Figure~\ref{fig:hypers_llama8} (a). 
Furthermore, we investigate the optimal number of insertion layers $n$ of \ourmethod. Section~\ref{sec:2} shows minimal divergence in shallow layers. Most LLMs have layer counts in multiples of four (e.g., 28, 32). We therefore position \ourmethod at three types of depths: the final 25\%, 50\%, and 75\% layers ($n$ = 25\%, 50\%, 75\%, respectively) of the LLM. Figure~\ref{fig:hypers_llama8} (b) demonstrates that best performance is achieved with \ourmethod at the final 50\% layers position. This could be because $n$ = 25\% inadequately improves information flows, while $n$ = 75\% causes misjudgment of critical content due to the slight divergence among information flows of different paragraphs in LLMs' shallow layers.
Finally, we discuss the most appropriate values for $\alpha$ in Equation~\eqref{eq:alpha} and $\beta$ in Equation~\eqref{eq:beta}, which are used to balance $v_m$ and $g_m$ and determine the minimum value of $w_m$, respectively. Figure~\ref{fig:hypers_llama8} (c)(d) demonstrate $\alpha$ = 1 and $\beta$ = 0.7 are the best choices.

\subsection{Further Analysis on Different Subsets} \label{sec:further}
\begin{table*}[t]
    \centering
    \caption{Results of original and shuffled input orderings when applying \ourmethod.}
    \label{tab:order}
    \resizebox{\textwidth}{!}{
    \begin{tabular}{c|c|cccccc|c} \toprule
    \multirow{2}{*}{Models} & \multirow{2}{*}{Variants} & \multirow{2}{*}{HotpotQA} & \multirow{2}{*}{2WikiMQA} & \multirow{2}{*}{MuSiQue} & \multicolumn{3}{c|}{LongBench} & \multirow{2}{*}{Average} \\ \cmidrule{6-8} 
    & & & & & HotpotQA & 2WikiMQA & MuSiQue & \\ \midrule
    \multirow{2}{*}{Qwen2.5-7B} 
    & Original & \bf 46.1 & \bf 49.9 & 35.0 & \bf 57.7 & 52.7 & \bf 33.4 & 45.8 \\
    & Shuffled & 46.1 & 49.7 & \bf 35.4 & 57.6 & \bf 53.0 & 33.3 & \bf 45.9 \\
    \midrule
    \multirow{2}{*}{Llama-3.1-8B} 
    & Original & 47.1 & 50.8 & \bf 39.2 & \bf 56.5 & 47.3 & \bf 32.0  & \bf 45.5 \\
    & Shuffled & \bf 47.2 & \bf 51.1 & 39.0 & 56.2 & \bf 47.5 & 31.8  & 45.5 \\
    \midrule
    \multirow{2}{*}{Qwen2.5-14B} 
    & Original & \bf 51.8 & 58.2 & \bf 43.8 & 60.9 & \bf 56.1 & \bf 39.3 & \bf 51.7 \\
    & Shuffled & 51.7 & \bf 58.3 & 43.8 & 60.9 & 56.0 & 39.3 & 51.7 \\
    \bottomrule
    \end{tabular}}
\end{table*}

\begin{table*}[t]
    \centering
    \small
    \caption{Results on different subsets of MuSiQue. \#Sup, \#T denote supporting paragraph count and total paragraph count, respectively.}
    \label{tab:count}
    \begin{tabular}{c|c|ccccccc} \toprule
    Models & Variants & \#Sup=2 & \#Sup=3 & \#Sup=4 & \#T=10 & \#T=20 & \#T=30 & \#T=40 \\ \midrule
    \multirow{2}{*}{Qwen2.5-7B} 
    & Vanilla & 30.4 & 30.9 & 30.3 & 32.3 & 30.5 & 29.8 & 29.4 \\
    & \ourmethod & \bf 35.0 & \bf 35.2 & \bf 34.7 & \bf 36.7 & \bf 35.0 & \bf 34.8 & \bf 34.3 \\
    \midrule
    \multirow{2}{*}{Llama-3.1-8B} 
    & Vanilla & 35.8 & 32.9 & 33.6 & 35.4 & 34.6 & 33.8 & 32.0 \\
    & \ourmethod & \bf 40.2 & \bf 35.7 & \bf 36.5 & \bf 39.5 & \bf 39.2 & \bf 38.7 & \bf 38.3 \\
    \midrule
    \multirow{2}{*}{Qwen2.5-14B} 
    & Vanilla & 38.1 & 37.3 & 38.7 & 39.3 & 38.0 & 36.9 & 36.1 \\
    & \ourmethod & \bf 44.0 & \bf 43.3 & \bf 44.3 & \bf 44.7 & \bf 43.8 & \bf 43.5 & \bf 42.9 \\
    \bottomrule
    \end{tabular}
\end{table*}
Since \ourmethod enhances attention on centrally located critical paragraphs, thereby mitigating the LLMs' inherent ``lost‑in‑the‑middle'' issue. It is essential to evaluate its robustness to different input orders. To this end, we conduct an experiment in which we randomize the ordering of paragraphs in all test samples while deliberately increasing the probability of supporting paragraphs occurring at the edges. This shuffled configuration intentionally weakens the attention of \ourmethod to these important information. As shown in Table \ref{tab:order}, results under both settings (using the fixed configuration $K$ = 10, $n$=50\%, $\alpha$ = 1, and $\beta$ = 0.7) reveal only minor performance fluctuations across ordering conditions. This stability demonstrates the robustness of \ourmethod to input order, primarily benefitting from our weighting strategy. Specifically, when evidence appears at the edges (where $g_m$ in Equation~\eqref{eq:gm} is reduced), \ourmethod attenuates the adjustments to the model's native attention distribution. This design intentionally leverages the model's inherent capacity to identify and amplify information flows from edge-positioned evidence, which aligns with the ``U-shaped attention bias'' phenomenon~\cite{ushape}.

In addition, we further analyze how paragraph counts affect \ourmethod. Since the MuSiQue dataset contains 2-4 supporting paragraphs per sample, we group results by supporting paragraph count and report the comparative results in Table~\ref{tab:count}. We observe minimal performance gaps across the groups, indicating that \ourmethod better captures relevant information flows to improve accuracy. To examine the effect of total paragraph count, we select all samples from MuSiQue and construct inputs with 10, 20, 30, and 40 paragraphs. The 20‑paragraph setting uses the original sample inputs; the 10‑paragraph setting randomly removes ten of the negative paragraphs; the 30‑paragraph and 40‑paragraph settings add 10/20 paragraphs sampled from other examples. Table~\ref{tab:count} shows that performance generally declines as the total paragraph count increases. However, \ourmethod degrades more slowly, since it suppresses information flow from negative paragraphs.

\section{Background and Related Works}
\textbf{Long-context Reasoning in LLMs.}
Recent advances in LLMs have spurred the development of long-context models (e.g., Llama-3.1-8B-Instruct~\cite{llama3} and Qwen2.5-7B-Instruct~\cite{qwen}, claiming 128k-token capacities). However, mainstream capacity benchmarks like Needle-in-a-Haystack (NIAH)~\cite{niah} mainly assess simple retrieval tasks, insufficiently reflecting complex real-world applications such as Multi-doc QA requiring cross-document information aggregation and multi-step reasoning. Current methods to enhance LLMs' long-context reasoning abilities follow two directions: (i) Positional encoding extensions (e.g., PI~\cite{pi}, CLEX~\cite{clex}, and CREAM~\cite{cream}) effectively expand context windows by adjusting position embeddings, but demand extra training resources and struggle to integrate with existing LLMs with inherently large context windows. (ii) Attention mechanism optimizations reduce computational costs and support longer inputs with minimal additional training. However, most approaches prioritize efficiency over actual reasoning improvements. Both strategies leave significant gaps in enhancing LLMs' reasoning abilities within their designated context windows.

\textbf{Attention mechanism.}
The Transformer~\cite{attention} architecture LLMs currently face two challenges: quadratic computational and memory scaling with input length and the ``lost-in-the-middle''~\cite{lost_in_the_middle} phenomenon. To overcome these constraints, recent studies are proposed to optimize the standard attention framework, including StreamingLLM~\cite{StreamingLLM}, LM-Infinite~\cite{lm_infinite}, H2O~\cite{h2o}, PINE~\cite{pine}, and SSC~\cite{ssc}.
StreamingLLM and LM-Infinite optimize memory usage by retaining initial tokens and recent tokens for next-token prediction, balancing efficiency and performance. H2O preserves recent tokens while dynamically selecting critical tokens (termed heavy hitters) to balance local relevance and global importance. PINE employs bidirectional inter-segment attention and re-assigns paragraph positions. SSC scales hidden states to enable more balanced attention distribution across different segments.
Collectively, current approaches either truncate long contexts to reduce computational and memory demands (e.g., StreamingLLM, LM-Infinite, H2O) or overlook the long-range modeling among paragraphs (e.g., PINE, SSC). Therefore, we propose \ourmethod, a training-free framework that adaptively optimizes attention matrices. Other approaches (e.g., prompt compression~\cite{prompt_compression}, retrieval-augmented generation~\cite{rag}, memory tree~\cite{memory_tree}) require external models or higher computation.

\section{Conclusion}
In this work, we address the critical challenges of limited long-range modeling and persistent ``lost-in-the-middle'' issue in Multi-doc QA for LLMs. Through a systematic analysis of information flow patterns from the paragraph disparity level and the answer quality level, our study reveals layer-wise aggregation patterns in distinct information flows and differences between good and bad reasoning instances. To resolve these issues, we propose Dual-Stage Adaptive Sharpening (\ourmethod), a training-free, plug-and-play mechanism that adaptively sharpens attention focus through two synergistic components: Contextual Gate Weighting (CGW) and Reciprocal Attention Suppression (RAS).

This work highlights the unexploited potential of attention optimization for LLMs. By adaptively sharpening attention score matrices, \ourmethod offers a practical solution for real-world applications requiring cross-document reasoning. Future directions include extending \ourmethod to diverse long-context scenarios, exploring its integration with retrieval-augmented generation frameworks (e.g., RAG \cite{rag}, GraphRAG \cite{graphrag}), and investigating related LLM applications (i.e., backdoor attacks \cite{attack}, data distillation \cite{beyond}, graph reasoning \cite{graphcogent}). Our findings underscore the importance of optimal attention optimization in unlocking the full capabilities of modern LLMs for complex tasks.

\section*{Acknowledgements}
This work is supported by National Natural Science Foundation of China No.62406057, No.62176046, the Fundamental Research Funds for the Central Universities No.ZYGX2025XJ042, the Noncommunicable Chronic Diseases-National Science and Technology Major Project No.2023ZD0501806, and the Sichuan Science and Technology Program under Grant No.2024ZDZX0011.

\bibliographystyle{plainnat}
\bibliography{neurips_2025}

%%%%%%%%%%%%%%%%%%%%%%%%%%%%%%%%%%%%%%%%%%%%%%%%%%%%%%%%%%%%
\newpage
\section*{NeurIPS Paper Checklist}

\begin{enumerate}

\item {\bf Claims}
    \item[] Question: Do the main claims made in the abstract and introduction accurately reflect the paper's contributions and scope?
    \item[] Answer: \answerYes{} % Replace by \answerYes{}, \answerNo{}, or \answerNA{}.
    \item[] Justification: The abstract and introduction clearly outline the core contributions of the paper: (1) systematic analysis of information flow in LLMs for Multi-doc QA tasks. (2) proposal of Dual-Stage Adaptive Sharpening (\ourmethod), a training-free plug-and-play framework with two modules: Contextual Gate Weighting (CGW) for alleviating ``lost-in-the-middle'' and Reciprocal Attention Suppression (RAS) for long-range dependency modeling optimization. (3) extensive experiments demonstrating DSAS's universality across diverse LLMs (average 4.2\% F1-score improvement on Llama-3.1-8B-Instruct and Qwen2.5-14B-Instruct). These align with the analysis (Section~\ref{sec:2}), methodology (Section~\ref{sec:method}) and experimental results (Section~\ref{sec:exp}).

\item {\bf Limitations}
    \item[] Question: Does the paper discuss the limitations of the work performed by the authors?
    \item[] Answer:  \answerYes{} % Replace by \answerYes{}, \answerNo{}, or \answerNA{}.
    \item[] Justification: Appendix~\ref{app:limit} explicitly discuss limitations, including DSAS's limited applicability to other long-context tasks and its high memory and computational costs when handling extremely lengthy inputs.

\item {\bf Theory assumptions and proofs}
    \item[] Question: For each theoretical result, does the paper provide the full set of assumptions and a complete (and correct) proof?
    \item[] Answer:  \answerYes{} % Replace by \answerYes{}, \answerNo{}, or \answerNA{}.
    \item[] Justification: Our framework's theoretical foundation about LLMs' information flow patterns is detailed in Section~\ref{sec:2}. Figures~\ref{fig:layer_ana},~\ref{fig:rea_ana} provide quantitative evidence of LLMs' information flow patterns.

\item {\bf Experimental result reproducibility}
    \item[] Question: Does the paper fully disclose all the information needed to reproduce the main experimental results of the paper to the extent that it affects the main claims and/or conclusions of the paper (regardless of whether the code and data are provided or not)?
    \item[] Answer:  \answerYes{} % Replace by \answerYes{}, \answerNo{}, or \answerNA{}.
    \item[] Justification: The paper provides comprehensive experimental details necessary for reproducibility. Section~\ref{sec:2} describes the experimental configuration for information flow analysis on the Multi-doc QA tasks, including dataset specifications and model implementations. Supplementary details such as prompt templates are provided in Appendix~\ref{app:templ}. In Section~\ref{sec:imp}, datasets (HotpotQA, 2WikiMultiHopQA, MuSiQue, LongBench), model architectures (Llama, Qwen, Mistral, Deepseek variants), and hyperparameter settings ($K$, $n$, $\alpha$, $\beta$) are explicitly defined. Furthermore, the implementations of ``bfloat16'' deployment and greedy-decoding are used to ensure consistency. 

\item {\bf Open access to data and code}
    \item[] Question: Does the paper provide open access to the data and code, with sufficient instructions to faithfully reproduce the main experimental results, as described in supplemental material?
    \item[] Answer:  \answerYes{} % Replace by \answerYes{}, \answerNo{}, or \answerNA{}.
    \item[] Justification: Datasets (HotpotQA, 2WikiMultiHopQA, MuSiQue and LongBench) are publicly available with citations provided in Section~\ref{sec:2} and Section~\ref{sec:imp}. License information for all datasets is included in references. Our code will be available once it is organized.

\item {\bf Experimental setting/details}
    \item[] Question: Does the paper specify all the training and test details (e.g., data splits, hyperparameters, how they were chosen, type of optimizer, etc.) necessary to understand the results?
    \item[] Answer: \answerYes{} % Replace by \answerYes{}, \answerNo{}, or \answerNA{}.
    \item[] Justification: Section~\ref{sec:2} and Section~\ref{sec:imp} describe datasets, models, hyperparameter settings ($K$, $n$, $\alpha$, $\beta$), and experimental environments. Appendix~\ref{app:templ} and Appendix~\ref{app:ans_ass} illustrate experimental templates and assessment method, respectively.

\item {\bf Experiment statistical significance}
    \item[] Question: Does the paper report error bars suitably and correctly defined or other appropriate information about the statistical significance of the experiments?
    \item[] Answer: \answerYes{} % Replace by \answerYes{}, \answerNo{}, or \answerNA{}.
    \item[] Justification: The information flow analysis in Section~\ref{sec:2} are conducted with 1000 test instances per task. To ensure the universality of our \ourmethod, we incorporate multiple LLMs (Llama, Qwen, etc.) in our experiments in Section~\ref{sec:exp}. Appendix~\ref{app:error} provides additional analysis of error cases.

\item {\bf Experiments compute resources}
    \item[] Question: For each experiment, does the paper provide sufficient information on the computer resources (type of compute workers, memory, time of execution) needed to reproduce the experiments?
    \item[] Answer: \answerYes{} % Replace by \answerYes{}, \answerNo{}, or \answerNA{}.
    \item[] Justification: The hardware specifications, including GPU type and memory, are detailed in Section~\ref{sec:imp}. Additionally, the computational complexity of our method is discussed in Appendix~\ref{app:complexity}.
    
\item {\bf Code of ethics}
    \item[] Question: Does the research conducted in the paper conform, in every respect, with the NeurIPS Code of Ethics \url{https://neurips.cc/public/EthicsGuidelines}?
    \item[] Answer: \answerYes{} % Replace by \answerYes{}, \answerNo{}, or \answerNA{}.
    \item[] Justification: The work focuses on attention optimization for Multi-doc QA tasks without ethical risks. All datasets are publicly available for academic use, adhering to their original licenses.

\item {\bf Broader impacts}
    \item[] Question: Does the paper discuss both potential positive societal impacts and negative societal impacts of the work performed?
    \item[] Answer: \answerNo{} % Replace by \answerYes{}, \answerNo{}, or \answerNA{}.
    \item[] Justification: The paper focuses on methodological contributions for attention optimization mechanism in the Multi-doc QA tasks. Broader impacts (e.g., misuse potential) are minimal as the work is foundational.
    
\item {\bf Safeguards}
    \item[] Question: Does the paper describe safeguards that have been put in place for responsible release of data or models that have a high risk for misuse (e.g., pretrained language models, image generators, or scraped datasets)?
    \item[] Answer: \answerNA{} % Replace by \answerYes{}, \answerNo{}, or \answerNA{}.
    \item[] Justification: This work uses publicly available LLMs (Llama, Qwen, etc.) and benchmark datasets (HotpotQA, 2WikiMultiHopQA, etc.). The paper does not introduce new pretrained language models, image generators, or scraped datasets that pose inherent misuse risks. All experiments involve standard academic datasets and LLMs cited in Section~\ref{sec:imp}, posing no special safety risks requiring safeguards.

\item {\bf Licenses for existing assets}
    \item[] Question: Are the creators or original owners of assets (e.g., code, data, models), used in the paper, properly credited and are the license and terms of use explicitly mentioned and properly respected?
    \item[] Answer: \answerYes{} % Replace by \answerYes{}, \answerNo{}, or \answerNA{}.
    \item[] Justification: All datasets (HotpotQA, 2WikiMultiHopQA, etc.) and LLMs (Llama, Qwen, etc.) are cited with original references (Section~\ref{sec:imp}). Publicly available datasets and LLMs are used under their respective licenses.

\item {\bf New assets}
    \item[] Question: Are new assets introduced in the paper well documented and is the documentation provided alongside the assets?
    \item[] Answer: \answerNA{} % Replace by \answerYes{}, \answerNo{}, or \answerNA{}.
    \item[] Justification: No new datasets, models, or codebases are released. The paper focuses on attention optimization methodology for enhanced Multi-QA performance on existing benchmarks (HotpotQA, 2WikiMultiHopQA, etc.). Experimental implementations will be shared post-acceptance following anonymity guidelines.

\item {\bf Crowdsourcing and research with human subjects}
    \item[] Question: For crowdsourcing experiments and research with human subjects, does the paper include the full text of instructions given to participants and screenshots, if applicable, as well as details about compensation (if any)? 
    \item[] Answer: \answerNA{} % Replace by \answerYes{}, \answerNo{}, or \answerNA{}.
    \item[] Justification: All experiments use publicly available Multi-doc QA datasets. No human subjects, crowdsourced data, or participant compensation are involved in this research.

\item {\bf Institutional review board (IRB) approvals or equivalent for research with human subjects}
    \item[] Question: Does the paper describe potential risks incurred by study participants, whether such risks were disclosed to the subjects, and whether Institutional Review Board (IRB) approvals (or an equivalent approval/review based on the requirements of your country or institution) were obtained?
    \item[] Answer: \answerNA{} % Replace by \answerYes{}, \answerNo{}, or \answerNA{}.
    \item[] Justification: This work does not involve human subjects, biological data, or sensitive personal information. The research exclusively analyzes anonymized text data from academic benchmarks, exempt from IRB oversight per institutional guidelines.

\item {\bf Declaration of LLM usage}
    \item[] Question: Does the paper describe the usage of LLMs if it is an important, original, or non-standard component of the core methods in this research? Note that if the LLM is used only for writing, editing, or formatting purposes and does not impact the core methodology, scientific rigorousness, or originality of the research, declaration is not required.
    %this research? 
    \item[] Answer: \answerYes{} % Replace by \answerYes{}, \answerNo{}, or \answerNA{}.
    \item[] Justification: The paper proposes \ourmethod, which operates as a plug-and-play attention optimization mechanism applied to existing Transformer-based LLMs for enhanced Multi-doc QA performance. The LLMs are the basis of the methodology. 

\end{enumerate}

%%%%%%%%%%%%%%%%%%%%%%%%%%%%%%%%%%%%%%%%%%%%%%%%%%%%%%%%%%%%
\newpage
\appendix
\section{Details of Information Flow Analysis on Multi-doc QA}
\subsection{Experimental Templates} \label{app:templ}
Consistent with LongBench~\cite{longbench}, the templates for HotpotQA~\cite{hotpotqa}, 2WikiMultiHopQA~\cite{2wikimqa}, and MuSiQue~\cite{musique} remain the same. Each data instance includes a question and its context, with task instructions positioned at both the start and end of the prompt to enhance model comprehension. This template is maintained consistently throughout our experiments in Section~\ref{sec:exp}.

\begin{tcolorbox}[size=title,breakable]
\noindent
Answer the question based on the given paragraphs. Only give me the answer and do not output any other words.\\ 
The following are given paragraphs.\\
\{context\}\\ 
Answer the question based on the given paragraphs. Only give me the answer and do not output any other words.\\ 
Question: \{question\}\\
Answer:
\end{tcolorbox}

\subsection{Paragraph Disparity Level Analysis on Qwen} \label{app:para_ana}
\begin{figure}[H]
    \centering
    \includegraphics[width=\linewidth]{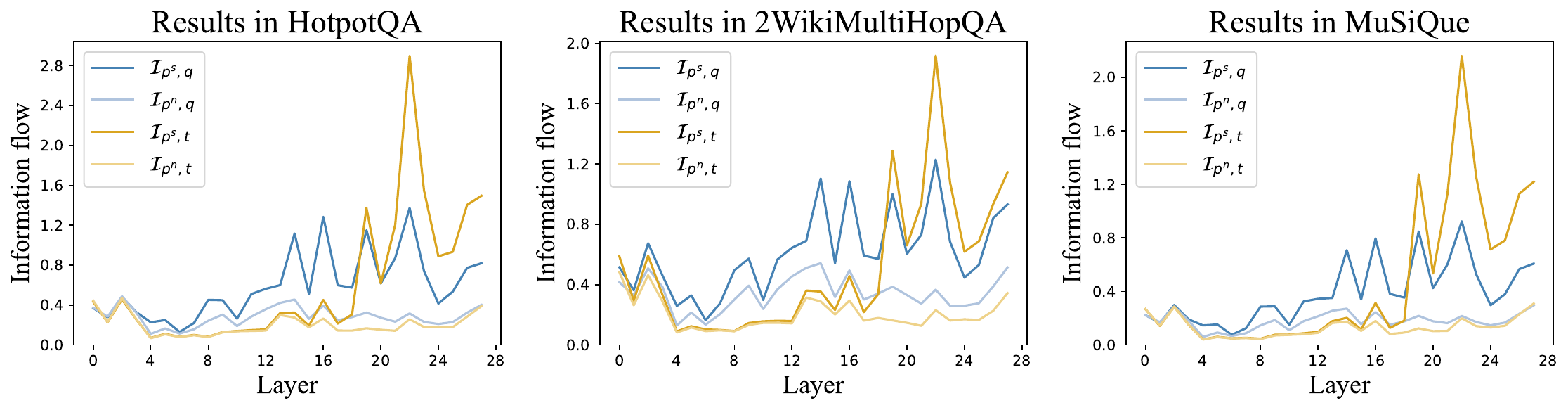}
    \vspace{-1em}
    \caption{Layer-wise information flows of HotpotQA, 2WikiMultiHopQA and MuSiQue tested on Qwen2.5-7B. $p^s$ and $p^n$ denote supporting paragraphs and negative paragraphs, respectively.}
    \label{fig:layer_qwen}
\end{figure}

Qwen2.5-7B consists of 28 stacked decoder layers. Analytical results on Figure~\ref{fig:layer_qwen} demonstrate similar conclusions to Llama-3.1-8B, with minimal information flow variations in shallow layers, followed by emerging divergences between $\mathcal{I}_{p^s,q}$ and $\mathcal{I}_{p^n,q}$, and finally gaps between $\mathcal{I}_{p^n,q}$ and $\mathcal{I}_{p^n,t}$. Notably, Qwen2.5-7B exhibits a slower progression of these divergences compared to Llama-3.1-8B, with the HotpotQA and the MuSiQue datasets particularly showing no significant information flow variations across the first 50\% of the model's layers.

\subsection{Answer Quality Level Analysis on Qwen} \label{app:ans_ana}
Similar to observations in Llama-3.1-8B, significant differences in information flow patterns emerge between good and bad reasoning instances, a consistent phenomenon observed in LLMs that forms the basis of \ourmethod framework.

\begin{figure}[H]
    \centering
    \includegraphics[width=.7\linewidth]{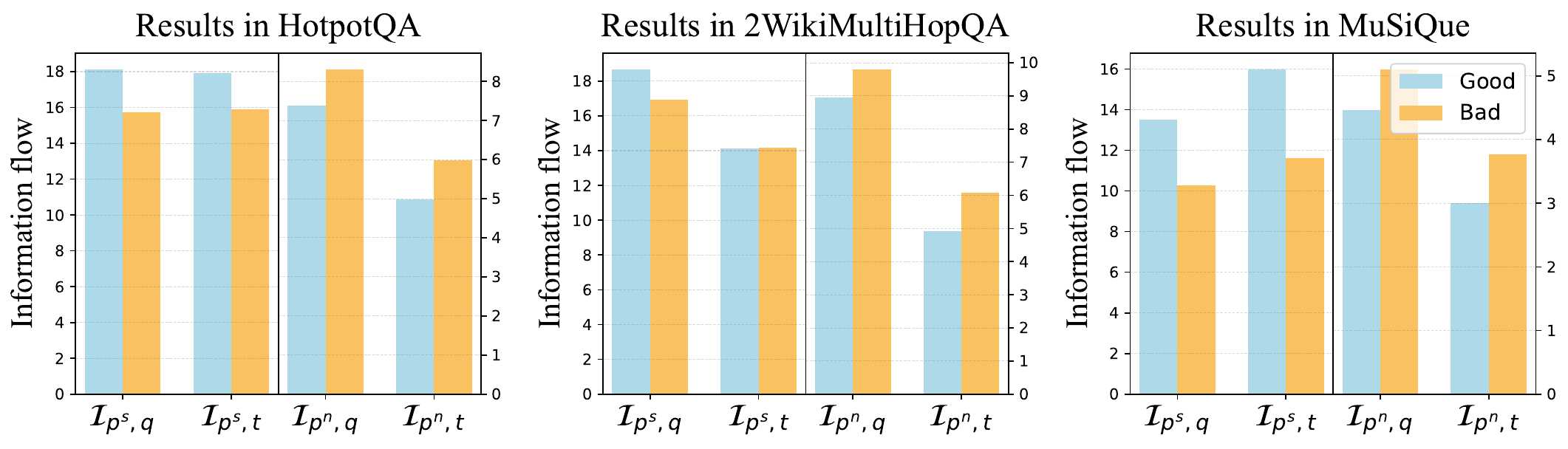}
    \caption{Comparison between mean values of the good and
        bad instances from the 1000 samples of Qwen2.5-7B. }
    \label{fig:bar_qwen}
\end{figure}

\subsection{Answer Quality Assessment} \label{app:ans_ass}
The evaluation method employed in these Multi-doc QA benchmarks adopts the answer quality assessment approach from HotpotQA. Specifically, for each data instance, both the reference answer and response first undergo normalization, including lowercase conversion of all textual content, elimination of non-essential articles (i.e., a/an/the), and redundant whitespace. Then precision and F1-score are computed between the normalized reference answer and response through exact lexical matching. This assessment ensures comparability across different benchmarks. We classify samples achieving a perfect F1-score (equal to 1) as \textsc{good} reasoning cases, and assign those with zero precision (equal to 0) to the \textsc{bad} reasoning instances. The assessment method remains the same for experiments at Section~\ref{sec:exp}.
\section{Symbol Explanation} \label{app:symb}
\begin{table}[h]
    \centering
    \caption{Symbols and Descriptions.}
    \begin{tabular}{l|l}
    \toprule
    Symbol & Description \\  \midrule
    $A^S_{h,l}$ & the attention score matrix of the $h$-th head in the $l$-th layer \\
    $A^W_{h,l}$ & the attention weight matrix of the $h$-th head in the $l$-th layer \\
    $\mathcal{I}_{{comb}^m}$ & the combined information flow for the $m$-th paragraph \\
    $\mu_I, \sigma_I$ & the mean and the standard deviation of $\mathcal{I}_{comb}$ \\
    $v_m$ & the normalized value of $\mathcal{I}_{{comb}^m}$ \\
    $\mu_p, \sigma_p$ & the mean and the standard deviation of sequence $\{0,1,\ldots,L-1\}$ \\
    $\gamma_m$ & the positional value of the $m$-th paragraph \\
    $g_m$ & the position-aware weight of the $m$-th paragraph \\
    $w_m^{'}, w_m$ & the temporary and the final contextual gate weight of the $m$-th paragraph \\
    $K$ & the number of tokens selected to compute $\mathcal{I}_{{comb}^m}$ \\
    $n$ & the proportion of layers inserting \ourmethod \\
    $\alpha, \beta$ & hyperparameters for computing $w_m$ \\
    \bottomrule
    \end{tabular}
    \label{tab:symbol}
\end{table}
\section{Full Hyperparameter Study} \label{app:hyper_full}
\begin{figure}[H]
\begin{minipage}{\linewidth}
\centering
\includegraphics[width=\linewidth]{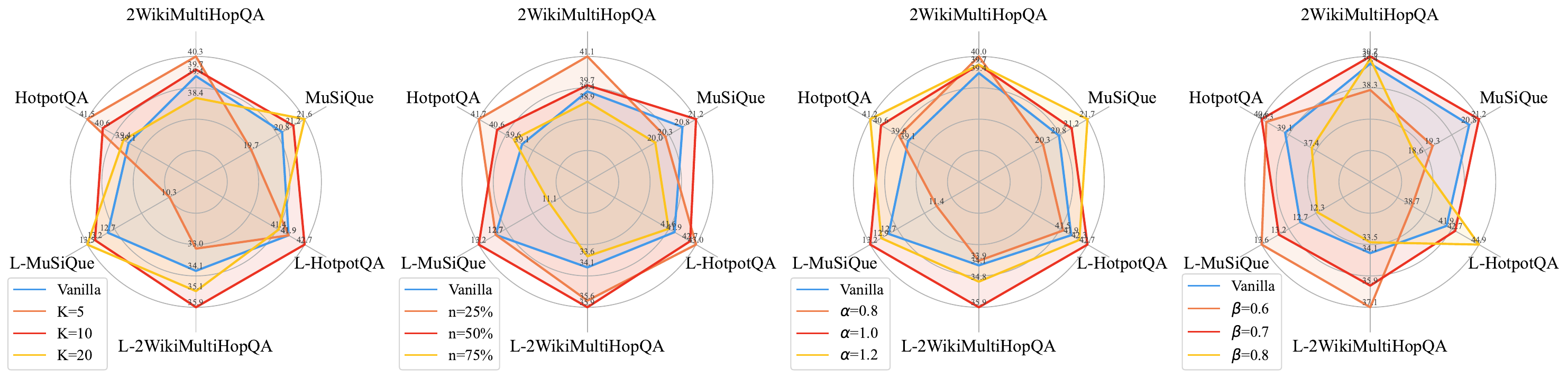}
\end{minipage}
\vfill
\centering
\vspace{-0.2em}
\includegraphics[width=\linewidth]{pics/legend.pdf}
\vspace{-1em}
\caption{Hyperparameter study of $K$, $n$, $\alpha$, $\beta$ on HotpotQA, 2WikiMultiHopQA, MuSiQue, and LongBench benchmarks of Llama-3.2-3B.}
\label{fig:hypers_llama3}
\end{figure}

\begin{figure}[H]
\begin{minipage}{\linewidth}
\centering
\includegraphics[width=\linewidth]{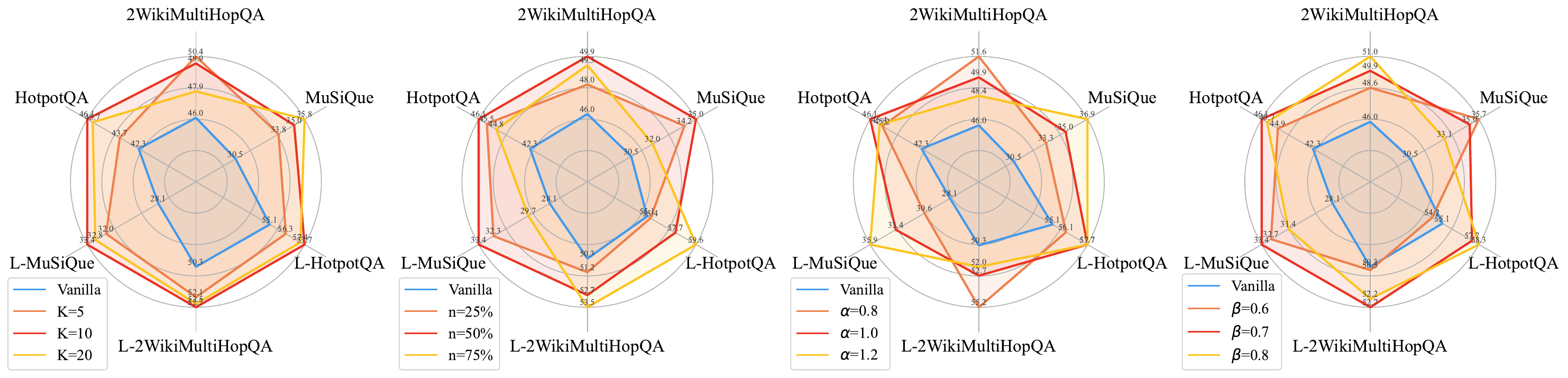}
\end{minipage}
\vfill
\centering
\vspace{-0.2em}
\includegraphics[width=\linewidth]{pics/legend.pdf}
\vspace{-1em}
\caption{Hyperparameter study of $K$, $n$, $\alpha$, $\beta$ on HotpotQA, 2WikiMultiHopQA, MuSiQue, and LongBench benchmarks of Qwen2.5-7B.}
\label{fig:hypers_qwen7}
\end{figure}

\begin{figure}[H]
\begin{minipage}{\linewidth}
\centering
\includegraphics[width=\linewidth]{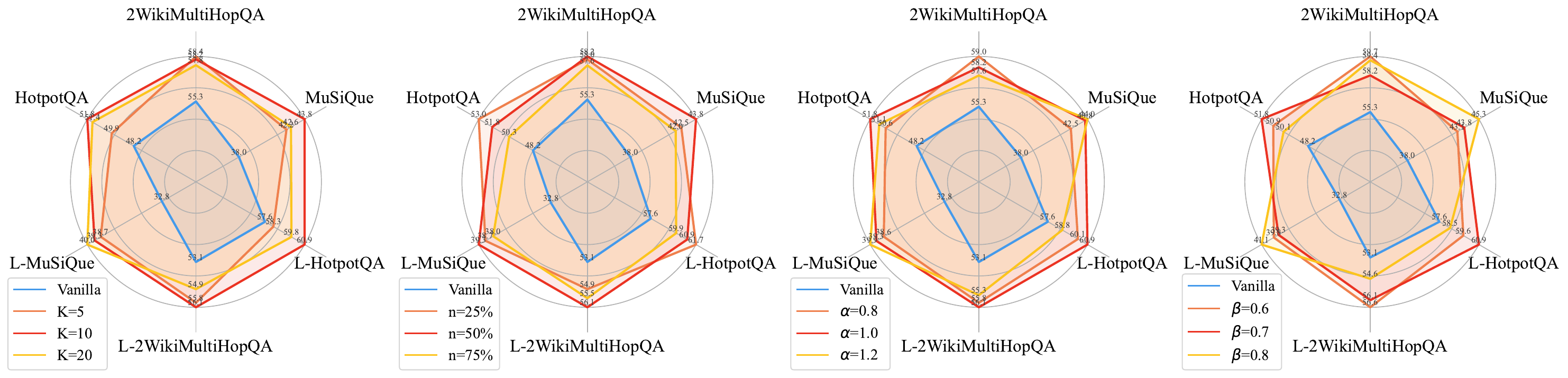}
\end{minipage}
\vfill
\centering
\vspace{-0.2em}
\includegraphics[width=\linewidth]{pics/legend.pdf}
\vspace{-1em}
\caption{Hyperparameter study of $K$, $n$, $\alpha$, $\beta$ on HotpotQA, 2WikiMultiHopQA, MuSiQue, and LongBench benchmarks of Qwen2.5-14B.}
\label{fig:hypers_qwen14}
\end{figure}

The results presented in Figures~\ref{fig:hypers_llama3},~\ref{fig:hypers_qwen7},~\ref{fig:hypers_qwen14} indicate that \ourmethod maintains consistent improvement across diverse LLMs under the hyperparameter configuration in Section~\ref{sec:imp} (red regions in the radar charts). Furthermore, we observe that the robustness of \ourmethod to hyperparameter variations strengthens with LLMs' scale expansion (3B, 7B, 14B), with Qwen2.5-14B particularly showing notable and stable performance improvements across all benchmarks under different hyperparameter settings. We believe that this phenomenon correlates with LLMs' scales and their intrinsic capabilities. Additionally, some hyperparameter configurations of \ourmethod may marginally degrade performance below baseline for Llama-3.2-3B, which further validates that our method achieves maximum performance gains specifically for middle-sized LLMs (7B, 8B, 14B).

\section{Error Analysis} \label{app:error}
We conduct an error analysis of \ourmethod with a focus on attention weight interactions among paragraphs (supporting and negative paragraphs), questions, and target. Following the experimental setup in Section~\ref{sec:2}, we randomly selected 1,000 samples for confusion matrix visualization between components. The analysis is conducted on the HotpotQA dataset due to its standardized structure of fixed 10-paragraph inputs (including 2 supporting paragraphs), which facilitates stable observation of interaction patterns.

\begin{figure}[H]
    \centering
    \includegraphics[width=\linewidth]{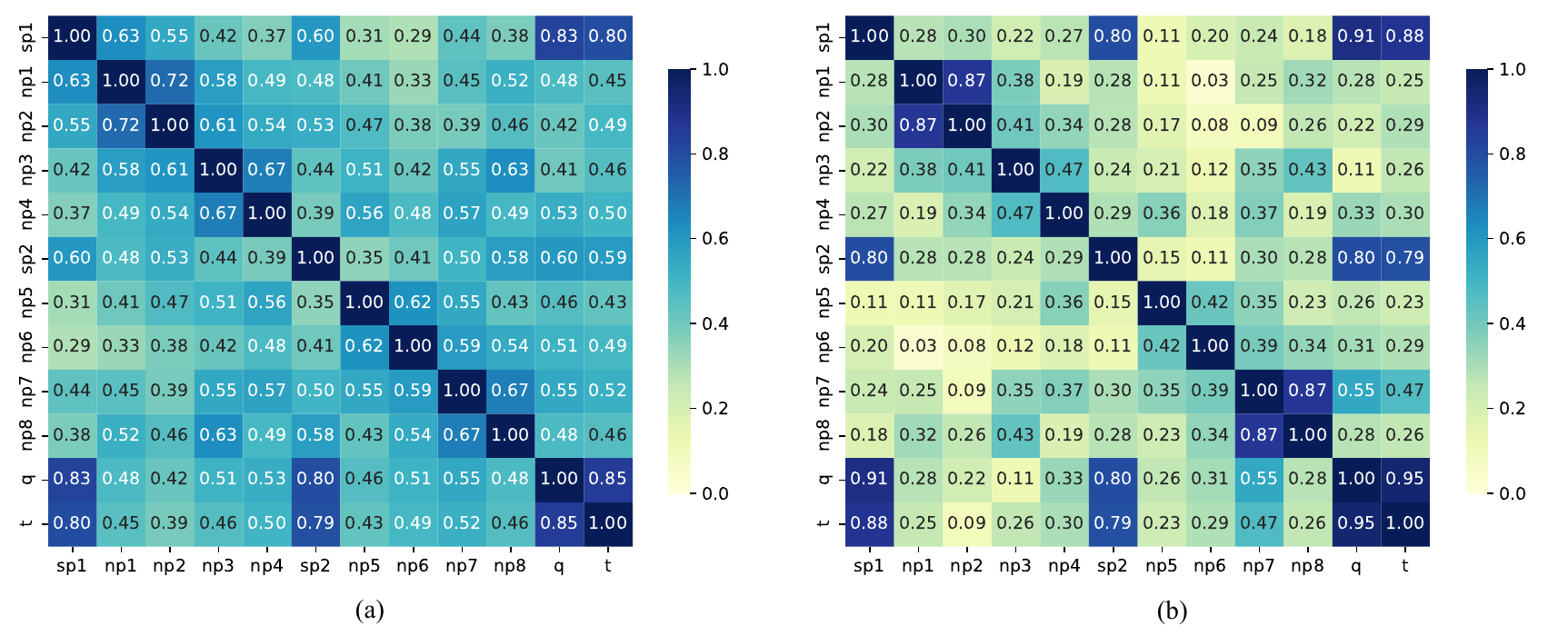}
    \caption{Confusion matrices on HotpotQA. (a) LLM, (b) LLM+DSAS. sp1, sp2 represent two supporting paragraphs. np1, np2, $\ldots$, np8 represent eight negative paragraphs. q and t denote question and target, respectively. The results are conducted on Llama-3.1-8B-Instruct.}
    \label{fig:error}
\end{figure}
Technically, we first average attention weights across all layers to obtain a global attention matrix. For pairwise component analysis (e.g., between two paragraphs), we extract corresponding sub-matrix based on token index ranges. All component pairs are analyzed following this method. Similar to Equation~\ref{eq:top_k}, we average the column-wise Top-K values for each sub-matrix. Finally we normalize all confusion values to 0-1. This approach effectively captures salient attention patterns while minimizing noise interference.
Figure~\ref{fig:error} presents heatmap comparisons of reasoning processes using LLM and LLM+DSAS. The results demonstrate that LLM+DSAS not only strengthens the focus of questions and targets on relevant support paragraphs, but also enhances information interactions between these supporting paragraphs. Additionally, it effectively suppresses unnecessary interactions between supporting paragraphs and irrelevant negative paragraphs.

\section{Complexity Analysis} \label{app:complexity}
Table~\ref{tab:complexity} presents the complexity of \ourmethod. Here, $L$ denotes the input token count, $Q$ denotes the question token count, and $C$ is the number of paragraphs. As illustrated in Equation~\ref{eq:CGW_RAS}, After computing the query, key, value matrices ($Q$, $K$, $V$) and attention scores $A_{h,l}^S$ through traditional attention mechanisms, we refine $A_{h,l}^S$ using the CGW and RAS modules. The updated scores are normalized via softmax to derive attention weights, which are then used to generate the final attention output. The core innovation of \ourmethod lies in computing contextual gate weight for dynamically weighting each paragraph. The time complexity scales linearly with $L$ and $Q$, contributing only a minor fraction of the total computation compared to the quadratic $\mathcal{O}(L^2)$ complexity of standard attention mechanisms. Furthermore, the space complexity grows linearly with $C$ alone. These characteristics ensure that \ourmethod maintains high efficiency in practice.
\begin{table*}[h]
    \centering
    \caption{Complexity analysis of \ourmethod.}
    \begin{tabular}{c|c}
    \toprule
        Time Complexity & $\mathcal{O}(LQ)$ \\
        Space Complexity & $\mathcal{O}(C)$ \\
    \bottomrule
    \end{tabular}
    \label{tab:complexity}
\end{table*}

\section{Scalability} \label{app:scale}
Evaluating the scalability of \ourmethod beyond multi-document QA presents an interesting research direction. To this end, we apply \ourmethod to several long-context tasks from LongBench~\cite{longbench}, including summarization (GovReport, QMSum, MultiNews) and code completion (LCC, RepoBench-P). The success of our information flow analysis largely relies on identifying anchors, which aggregate critical information from redundant contexts to guide answer generation. While in Multi-doc QA, the question and target naturally serve as anchors, extending \ourmethod to broader long-context tasks requires defining appropriate anchors for each scenario. After reviewing the prompt templates, we select the final instruction sentence or the query and the final sentence as anchors for aggregating information flows (e.g., ``Now, write a one‑page summary of the report.\textbackslash n Summary:'' for GovReport; ``Next line of code:'' for LCC).

\begin{table*}[h]
    \centering
    \small
    \caption{Comparison results on summarization and code completion tasks with hyperparameter configuration ``$K$=10, $n$=50\%, $\alpha$=1, $\beta$=0.7''. The evaluation metrics for L-GovReport, L-QMSum, L-MultiNews and L-LCC, L-RBP are Rouge-L score and Edit Sim, respectively. RBP denotes RepoBench-P.}
    \begin{tabular}{c|c|cccccc} \toprule
    Models & Variants & L-GovReport	&L-QMSum &L-MultiNews &L-LCC &L-RBP &Average \\ \midrule
    \multirow{2}{*}{Qwen2.5-7B} 
    & Vanilla & 33.6 &22.4 &23.7 &53.7 &48.2 &36.3 \\
    & \ourmethod &\bf37.0 &\bf25.1 &\bf26.2 &\bf56.3 &\bf51.5 &\bf39.2 \\
    \midrule
    \multirow{2}{*}{Llama-3.1-8B} 
    & Vanilla &34.9 &24.8 &27.1 &58.1 &50.8 &39.1\\
    & \ourmethod &\bf37.4 &\bf26.6 &\bf29.3 &\bf60.0 &\bf52.5 &\bf41.2 \\
    \midrule
    \multirow{2}{*}{Qwen2.5-14B} 
    & Vanilla &38.4 &28.2 &31.8 &63.9 &55.6 &43.6 \\
    & \ourmethod &\bf41.0 &\bf30.4 &\bf34.6 &\bf66.7 &\bf58.9 &\bf46.3 \\
    \bottomrule
    \end{tabular}
    \label{tab:scale}
\end{table*}

Since \ourmethod operates on paragraph-level segments, which Multi-doc QA inherently provides, we employ fixed-length token segmentation for the other tasks. Specifically, we use 500‑token segments for GovReport, QMSum, and RepoBench‑P, and 200‑token segments for MultiNews and LCC, ensuring a medium number (5-40) of paragraphs per example. The comparison results are presented in Table \ref{tab:scale}.
Based on our experimental results, we make two key observations: (i) \ourmethod consistently enhances performance across all five tasks under three medium-sized model architectures; and (ii) these improvements generalize to diverse tasks (summarization and code completion), demonstrating that \ourmethod is effective beyond multi-doc QA.

The prompt templates used for these tasks align with those in the LongBench~\cite{longbench} benchmark. The sentences in red in the following template boxes serve as anchors.

The template for L-GovReport is shown below:
\begin{tcolorbox}[size=title,breakable]
\noindent
You are given a report by a government agency. Write a one-page summary of the report.\\
Report:\\
\{context\}\\ 
\textcolor{red}{Now, write a one-page summary of the report.}\\
\textcolor{red}{Summary:}
\end{tcolorbox}
The template for L-QMSum is shown below:
\begin{tcolorbox}[size=title,breakable]
\noindent
You are given a meeting transcript and a query containing a question or instruction. Answer the query in one or more sentences.\\
Transcript:\\
\{context\}\\ 
Now, answer the query based on the above meeting transcript in one or more sentences. \\
\textcolor{red}{Query: \{input\}} \\
\textcolor{red}{Answer:} 
\end{tcolorbox}
The template for L-MultiNews is shown below:
\begin{tcolorbox}[size=title,breakable]
\noindent
You are given several news passages. Write a one-page summary of all news. \\ 
News:\\
\{context\}\\ 
\textcolor{red}{Now, write a one-page summary of all the news.} \\
\textcolor{red}{Summary:}
\end{tcolorbox}
The template for L-LCC is shown below:
\begin{tcolorbox}[size=title,breakable]
\noindent
Please complete the code given below. \\ 
\{context\}\\ 
\textcolor{red}{Next line of code:}
\end{tcolorbox}
The template for L-RepoBench-P is shown below:
\begin{tcolorbox}[size=title,breakable]
\noindent
Please complete the code given below. \\ 
\{context\}\{input\}\\ 
\textcolor{red}{Next line of code:}
\end{tcolorbox}

\section{Limitations} \label{app:limit}
While \ourmethod acts as a plug-and-play attention mechanism and consistently improves performance in Multi-doc QA tasks, our work has two primary limitations that warrant discussion. 
(1) The scalability of \ourmethod warrants further investigation. While our experiments on summarization and code completion tasks demonstrate consistent performance gains, the current fixed-token-count chunking strategy remains relatively simple. Future work could explore more refined and semantically-aware chunking methods for these tasks, which may further enhance information flow between key semantic segments and the generated answer.
(2) Although \ourmethod enhances performance within standard LLM context windows, it inherits the fundamental computational and memory limitations of conventional Transformer~\cite{attention} architectures, with quadratic complexity in both memory usage and computational cost. Consequently, our approach does not address the scalability challenges associated with processing extremely long documents (e.g., those exceeding 100K tokens), where memory and computational demands become prohibitive. Future investigations could consider integrating sparse attention strategies or memory-efficient architectures to mitigate these constraints.

These limitations highlight important directions for future research while not diminishing the effectiveness of \ourmethod. The existing framework establishes a reliable base for advancing attention optimization in long-context processing.

\end{document}